\documentclass[10pt,twocolumn,letterpaper]{article}

\usepackage{wacv}
\usepackage{times}
\usepackage{epsfig}
\usepackage{graphicx}
\usepackage{amsmath}
\usepackage{amssymb}
\usepackage{booktabs}
\usepackage{amsthm}
\usepackage{bbold}
\usepackage{multirow}

\usepackage{caption}
\usepackage{subcaption}

\usepackage[accsupp]{axessibility}

\newcommand{\norm}[1]{\left\lVert#1\right\rVert}

\def\wacvPaperID{770} 

\wacvalgorithmstrack   

\wacvfinalcopy 

\def\D{\mathcal{D}}

\def\R{\mathbb{R}}

\newcommand{\lossarg}[2]{\mathcal{#1}_{\mbox{\scriptsize #2}}}

\ifwacvfinal
\usepackage[breaklinks=true,bookmarks=false]{hyperref}
\else
\usepackage[pagebackref=true,breaklinks=true,colorlinks,bookmarks=false]{hyperref}
\fi
\usepackage{hyperref}

\usepackage[capitalize]{cleveref}
\crefname{section}{Sec.}{Secs.}
\Crefname{section}{Section}{Sections}
\Crefname{table}{Table}{Tables}
\crefname{table}{Tab.}{Tabs.}
\Crefname{appsec}{Appendix}{Appendices}
\crefname{appsec}{App.}{Apps.}

\def\our{{ProtoSeg}}

\begin{document}

\title{\our{}: Interpretable Semantic Segmentation with Prototypical Parts}

\author{Mikołaj Sacha$^1$
\and
Dawid Rymarczyk$^{1,2}$
\and
Łukasz Struski$^1$
\and
Jacek Tabor$^1$
\and
Bartosz Zieliński$^{1,3}$\\
$^1$ Faculty of Mathematics and Computer Science, Jagiellonian University, Kraków, Poland\\
$^2$ Ardigen, Kraków, Poland
$^3$ IDEAS NCBR, Warsaw, Poland \\
{\tt\small \{mikolaj.sacha;dawid.rymarczyk\}@doctoral.uj.edu.pl}\\
{\tt\small \{lukasz.struski;jacek.tabor;bartosz.zielinski\}@uj.edu.pl}
}

\maketitle
\thispagestyle{empty}

\begin{abstract}
We introduce \our{}, a novel model for interpretable semantic image segmentation, which constructs its predictions using similar patches from the training set. To achieve accuracy comparable to baseline methods, we adapt the mechanism of prototypical parts and introduce a diversity loss function that increases the variety of prototypes within each class. We show that \our{} discovers semantic concepts, in contrast to standard segmentation models. Experiments conducted on Pascal VOC and Cityscapes datasets confirm the precision and transparency of the presented method.
\end{abstract}

\section{Introduction}


Semantic segmentation is an essential component in many visual understanding systems. However, while deep learning-based models have achieved promising performance on challenging benchmarks~\cite{minaee2021image}, their decisions remain unclear due to lack of explanation~\cite{rudin2019stop}. This issue may appear particularly problematic in critical applications, such as medical imaging or autonomous driving.

\begin{figure}[t]
\begin{center}
\includegraphics[width=0.9\columnwidth]{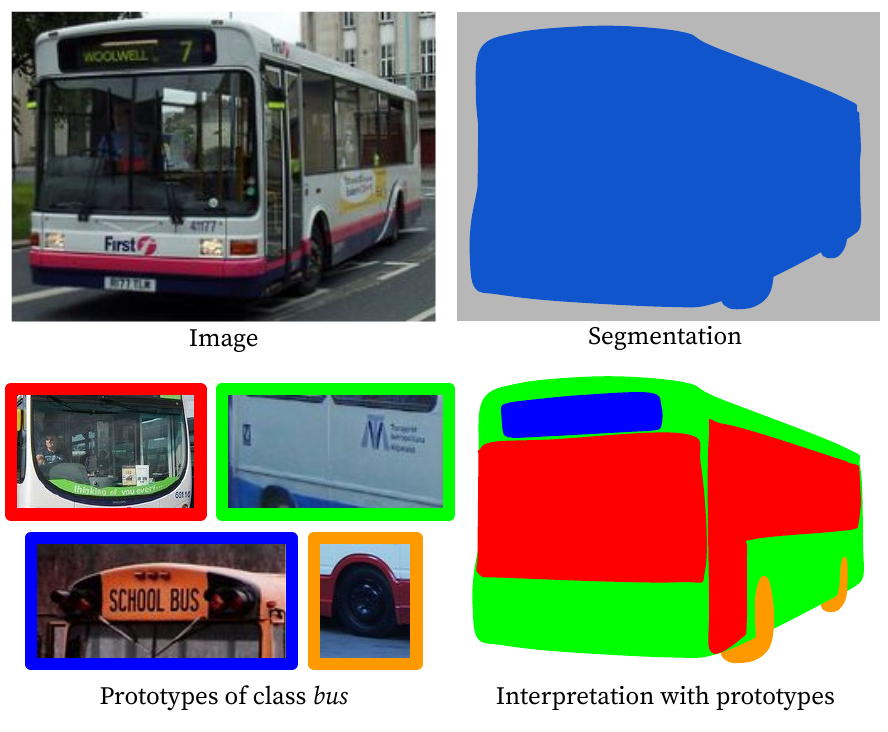}
\end{center}
\caption{In contrast to existing methods, \our{} provides an interpretation of resulted segmentation. For this purpose, it operates on patches selected from a training set (prototypes) corresponding to parts of the segmented objects. For a bus, prototypes can correspond to windows or wheels, represented by red and orange colors, respectively.}
\label{fig:object_parts_with_prototypes}
\end{figure}

Most of the eXplainable Artificial Intelligence (XAI) approaches focus on classification or regression task~\cite{basaj2021explaining,chen2019looks,gee2019explaining,selvaraju2017grad,zhang2021protgnn}. Therefore, interpretable segmentation is still considered an open question~\cite{rudin2022interpretable}, with only a few initial works on the crossroad of XAI and segmentation. One of them is the Symbolic Semantic (S2) framework~\cite{santamaria2020towards} where together with segmentation, the model generates a symbolic sentence derived from a categorical distribution. Another approach~\cite{vinogradova2020towards} generalizes the Grad-CAM method~\cite{selvaraju2017grad} to the problem of segmentation. However, both methods have significant disadvantages. The former requires a predefined vocabulary of symbolic words, while the latter can be unreliable and introduce additional bias to the results~\cite{NEURIPS2018_294a8ed2}.

\begin{figure*}[!ht]
\begin{center}
\includegraphics[width=0.9\linewidth]{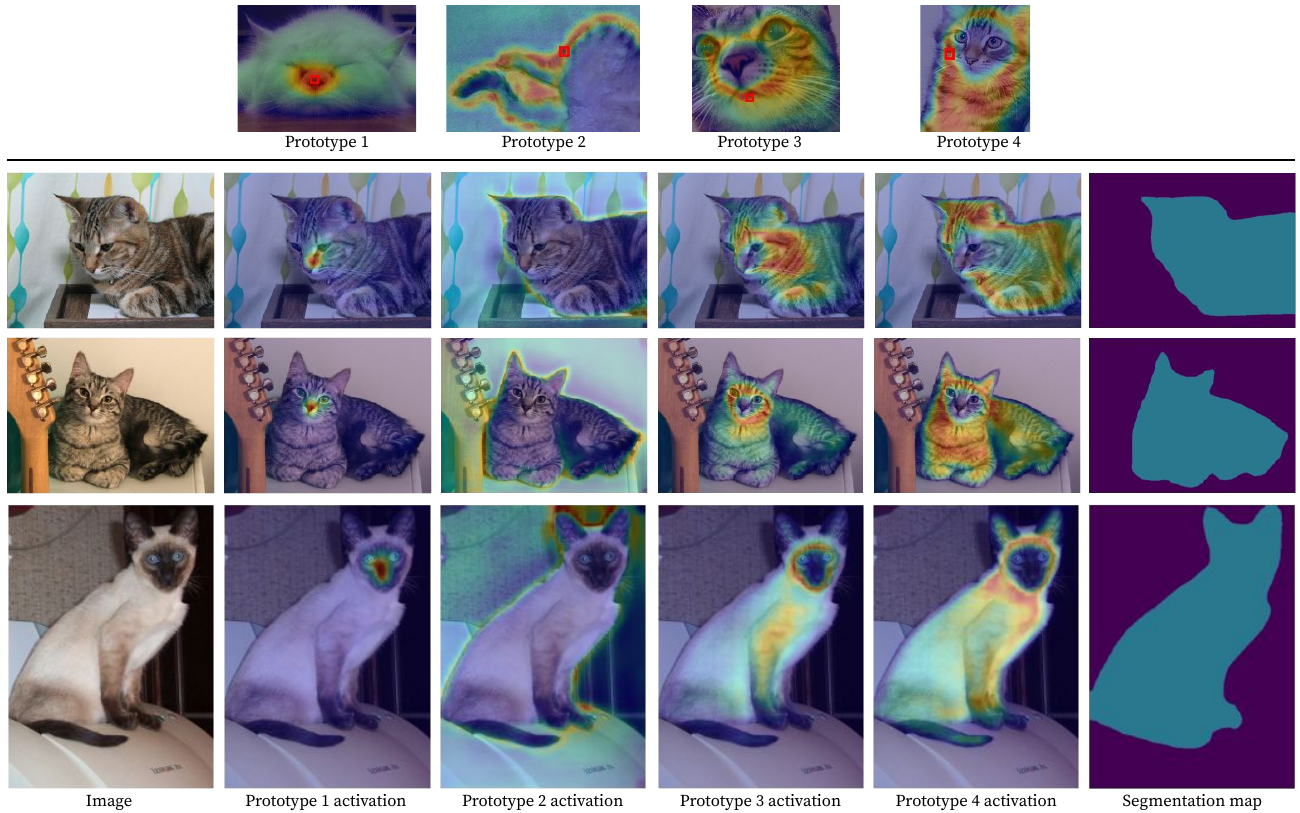}
\end{center}
\caption{Prototype activation maps generated by \our{} for four prototypes from class \textit{cat} (columns) and three sample images from PASCAL VOC 2012 (rows). Maps differ from each other, e.g. prototype 1 concentrates on the cat's nose, while prototype 4 activates mostly on the cat's neck. We see that \our{} can derive semantic concepts using prototypical cases from the training dataset.}
\label{fig:prototype_object_parts_cats}
\end{figure*}

In this paper, we introduce \our{}, an interpretable semantic segmentation method based on prototypical parts~\cite{chen2019looks}. While the standard approaches return only the class probability for each input pixel, \our{} learns prototypes for each class and uses them to generate and explain segmentation with patches (cases) from the training set. As we present in~\Cref{fig:object_parts_with_prototypes} and~\Cref{sec:results}, the main goal is to focus prototypes of the same class on different semantic concepts. For this purpose, we introduce a novel diversity loss function that increases the variety of prototypes for each class (see Figure~\ref{fig:prototype_object_parts_cats}).
Such application of case-based methodology significantly increases the interpretability of the segmentation model. Moreover, in contrast to previous methods, it does not require additional effort from the users to provide explanations.

To show the effectiveness of \our{}, we conduct experiments on three datasets: Pascal VOC 2021~\cite{pascal2012}, Cityscapes~\cite{cityscapes2016} and EM Segmentation Challenge~\cite{em_challenge}. The results indicate no significant decrease in performance between our interpretable model and the original black-box approaches like DeepLabv2~\cite{chen2017deeplab} or U-Net~\cite{ronneberger2015u}. Additionally, we present ablation studies showing how diversity loss influences model performance and transparency. We made the code available.
Our contributions can be therefore summarized as follows:
\begin{itemize}
    \item we introduce a model that employs prototypical parts to provide interpretable semantic segmentation,
    \item we define a diversity loss function based on Jeffrey's divergence that increases the variability of prototypes within each class.
    \item we show that \our{} can be used with different backbone architectures and on various semantic segmentation tasks.
\end{itemize}

In the next section of this paper, we discuss related works, then in \Cref{sec:protoseg} we introduce \our{} and the diversity loss function. Later, in \Cref{sec:exp_set}, we describe the experimental setup that uses DeepLab~\cite{chen2017deeplab} model as a backbone, followed by the results in \Cref{sec:results}, in which we also present the extendability of \our{} to different segmentation architectures. Finally, we conclude our work in \Cref{sec:conc}.

\begin{figure*}[th]
\begin{center}
\includegraphics[width=0.9\linewidth]{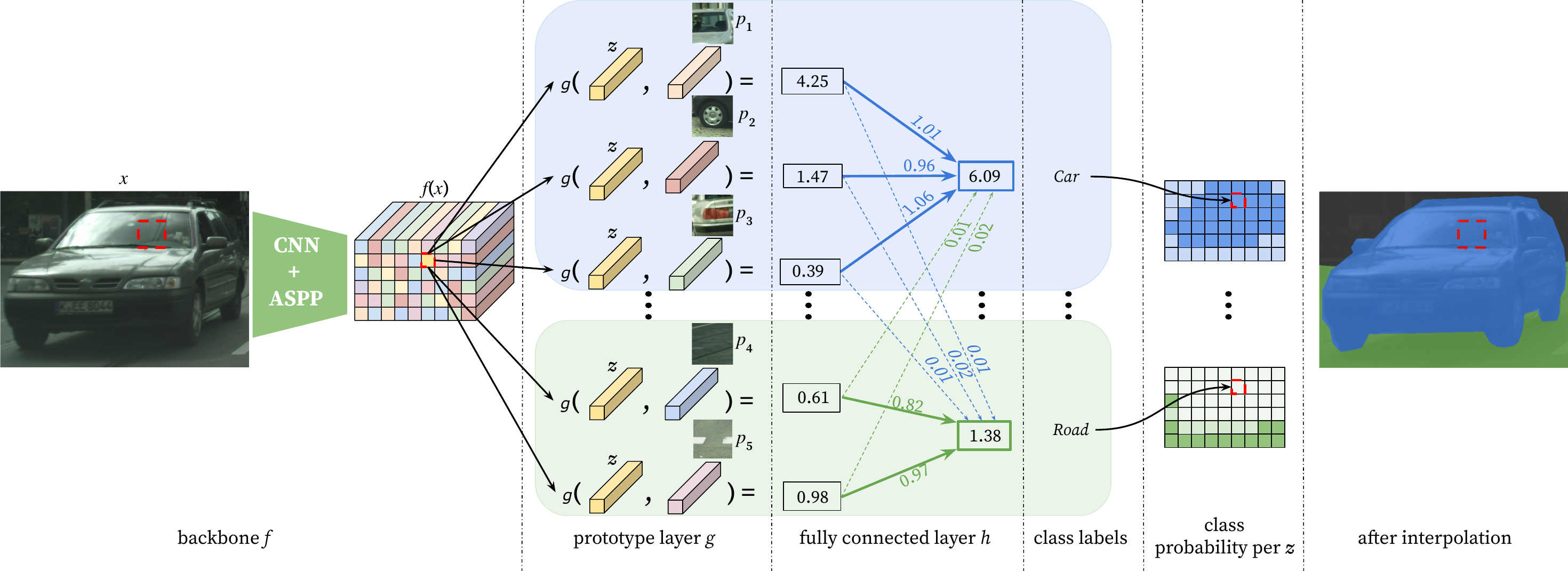}
\end{center}
\caption{\our{} consists of a backbone network $f$, prototype layer $g$, and a fully connected layer $h$. While the backbone network processes the image as a whole, the prototype and fully connected layers consider each $z$ from feature map $f(x)$ separately. The final segmentation is obtained by interpolating the output map corresponding to class probability.}
\label{fig:architecture}
\end{figure*}

\section{Related works}

\paragraph{Explainable artificial intelligence.}
Deep learning explanations can be obtained with two types of models: post hoc or self-explainable~\cite{rudin2019stop}. Post hoc approaches explain the reasoning process of black-box methods. They include a saliency map~\cite{marcos2019semantically,rebuffi2020there,selvaraju2017grad,selvaraju2019taking,simonyan2014deep} that is a heatmap of essential image parts, Concept Activation Vectors (CAV) revealing the internal network state as user-friendly concepts~\cite{chen2020concept,NEURIPS2019_77d2afcb,kim2018interpretability,pmlr-v119-koh20a,NEURIPS2020_ecb287ff}, counterfactual examples~\cite{abbasnejad2020counterfactual,goyal2019counterfactual,mothilal2020explaining,niu2021counterfactual,wang2020scout}, or analyzing the networks' reaction to the image perturbation~\cite{basaj2021explaining,fong2019understanding,fong2017interpretable,ribeiro2016should}. Post hoc methods are convenient because they do not require any changes to the models' architecture, but they may produce biased and fragile explanations~\cite{NEURIPS2018_294a8ed2}. For this reason, self-explainable models attract attention~\cite{NEURIPS2018_3e9f0fc9,brendel2018approximating} making the decision process more transparent.
Recently, many researchers have focused on enhancing the concept of the prototypical parts introduced in ProtoPNet~\cite{chen2019looks} to represent the networks' activation patterns. The most prominent extensions include TesNet~\cite{wang2021interpretable} and Deformable ProtoPNet~\cite{donnelly2022deformable} that exploits orthogonality in the prototype construction. ProtoPShare~\cite{rymarczyk2021protopshare}, ProtoTree~\cite{nauta2021neural}, and ProtoPool~\cite{rymarczyk2021interpretable} reduce the amount of prototypes used in the classification. Other methods consider hierarchical classification with prototypes~\cite{hase2019interpretable}, prototypical part transformation from the latent space to data space~\cite{li2018deep}, and knowledge distillation technique from prototypes~\cite{keswani2022proto2proto}. Moreover, prototype-based solutions are widely adopted in various applications, such as medical imaging~\cite{afnan2021interpretable,barnett2021case,kim2021xprotonet,rymarczyk2021protomil,singh2021these}, time-series analysis~\cite{gee2019explaining}, graphs classification~\cite{zhang2021protgnn}, and sequence learning~\cite{ming2019interpretable}. In this paper, we adapt the prototype mechanism to the semantic segmentation task.

\paragraph{Semantic segmentation.}
Similarly to other tasks of computer vision, recent semantic segmentation methods base on deep architectures~\cite{girshick2014rich,gupta2014learning,long2015fully}, especially convolutional neural networks~\cite{chen2014semantic,he2015spatial,liu2015parsenet,zhao2017pyramid}. Moreover, they usually consist of two parts: an encoder pretrained on a classification task and a decoder network semantically projecting the activation features onto the pixel space. The most popular models include U-Net~\cite{ronneberger2015u}, which contain shortcuts between the down-sampling layer in the encoder and the corresponding up-sampling layer in the decoder that effectively capture fine-grained information. Several works use additional mechanisms (such as conditional random fields) at the network output to improve models' performance~\cite{arnab2016bottom,chandra2016fast,lin2016efficient}. Some approaches adapt superpixels~\cite{mostajabi2015feedforward,sharma2015deep}, Markov random field~\cite{liu2015semantic}, or modules learning pixel affinities~\cite{liu2017learning,wang2018non} to obtain segmentation. Others employ contrastive learning~\cite{chen2021semi,hu2021region,zheng2021rethinking} or multiple receptive fields~\cite{yuan2020multi} to increase the segmentation quality. Moreover, with recent advancements in transformers architectures, models such as~\cite{gu2022multi,strudel2021segmenter,xie2021segformer,zheng2021rethinking} are used to obtain state-of-the-art results. Finally, Chen et al.~\cite{chen2017deeplab} proposed DeepLab method that uses multiple techniques to improve the existing methods: atrous convolutions, atrous spatial pyramidal pooling, and conditional random fields. We provide an interpretable version of this method.

\section{\our{}}
\label{sec:protoseg}

In this section, we first describe the architecture of our \our{} method for interpretable semantic segmentation. Then, we provide information about the training procedure. Finally, we describe a novel regularization technique that increases the variety of prototypes within each class.

\subsection{Architecture}
\label{sec:architecture}

\Cref{fig:architecture} illustrates the architecture of \our{}, composed of a backbone network $f$, prototype layer $g$, and a fully connected layer $h$.
Let $x \in \mathbb{R}^{H \times W \times 3}$ be an RGB image and feature map $f(x) \in \mathbb{R}^{H_d \times W_d \times D}$ be the output of the backbone network for this image. Moreover, let us consider $z \in \mathbb{R}^D$ as a point (or \textit{patch}) from $f(x)$. Each $z$ is passed to prototype layer $g$ with $M$ learnable prototypes $p_j \in \mathbb{R}^D$ to compute $M$ similarity scores (prototype's activations) using formula from~\cite{chen2019looks}:
\begin{equation}
g(z, p_j) = log\left(\frac{\norm{z-p_j}_2^2+1}{\norm{z-p_j}_2^2+\epsilon}\right).
\end{equation}
The $M$ similarity scores computed for feature map point $z$ are processed through a fully connected layer $h$ with weight matrix $w_h \in \mathbb{R}^{M \times C}$ to produce probabilities of $C$ classes. As a result of processing all $z$ from $f(x)$ through $g$ and $w_h$, we acquire output map of shape $H_d \times W_d \times C$. To obtain the final segmentation, this map is interpolated to resolution $H \times W \times C$.

As the backbone network, we use DeepLab~\cite{chen2017deeplab}, a standard model for image segmentation that consists of ResNet-101~\cite{he2015resnet} pretrained on some large-scale computer vision task, followed by an Atrous Spatial Pyramid Pooling (ASPP) layer. In the prototype layer $g$, as in~\cite{chen2019looks}, each prototype is assigned to one of the $C$ classes. We define $\mathbf{P}_c$ as the set of all prototypes from class $c \in C$ and initialize $w_h^{(c, j)}=1$ for all $p_j \in \mathbf{P}_c$ and $w_h^{(c, j)}=-0.5$ for all $p_j \notin \mathbf{P}_c$. This initialization steers the model towards producing high activation between feature map points and prototypes of their predicted class, while lowering their activation to prototypes from other classes.
Similarly to~\cite{chen2017deeplab}, the model's output is obtained differently in the inference and training phases. In inference, we use bilinear interpolation to match the size of the segmentation map with the input image size, while in training, we decrease the resolution of ground truth segmentation to fit the size of the output feature map.

\subsection{Multi-step training procedure}
\label{sec:multi_step_training}

We apply a multi-step training protocol from~\cite{chen2019looks}. We start from the ResNet-101 layers pretrained on same large visual recognition task, while randomly initializing ASPP and prototype vectors and setting the weights $w_h$ as described in~\Cref{sec:architecture}. We start with a warmup phase, where we freeze ResNet-101 and $w_h$ weights, training only ASPP and prototype layer. Then, we run a joint optimization process, where we train everything except $w_h$ weights. Next, we execute prototype projection, which replaces prototypes with the representation of the nearest patch $z$ from the training set. During this stage, we also remove duplicate prototypes that are projected onto same training patch. Then, in the tuning phase, we fine-tune the $w_h$ weights of the last layer. Subsequently, we apply prototype pruning that removes non-class-specific prototypes using the pruning algorithm from~\cite{chen2019looks}. Finally, we again fine-tune the last layer $h$.

\subsection{Diverse prototypes of same class}

\begin{figure}[ht]
\centering
\begin{subfigure}[b]{\columnwidth}
\centering
\includegraphics[width=\columnwidth]{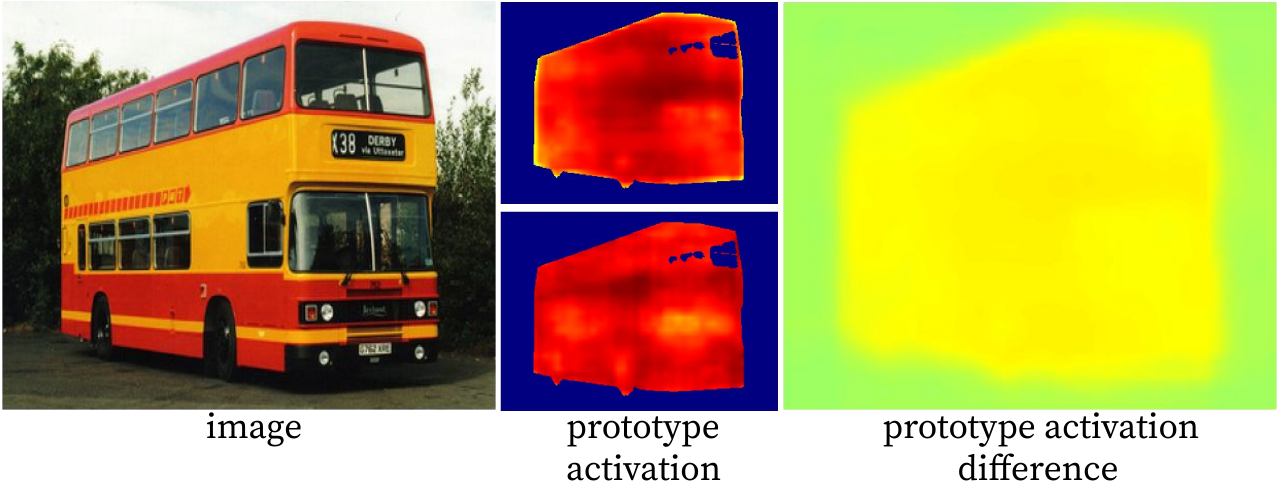}
\vspace*{-5mm}
\caption{High value of $\lossarg{L}{J}$.}
\vspace*{4mm}
\end{subfigure}
\begin{subfigure}[b]{\columnwidth}
\centering
\includegraphics[width=\columnwidth]{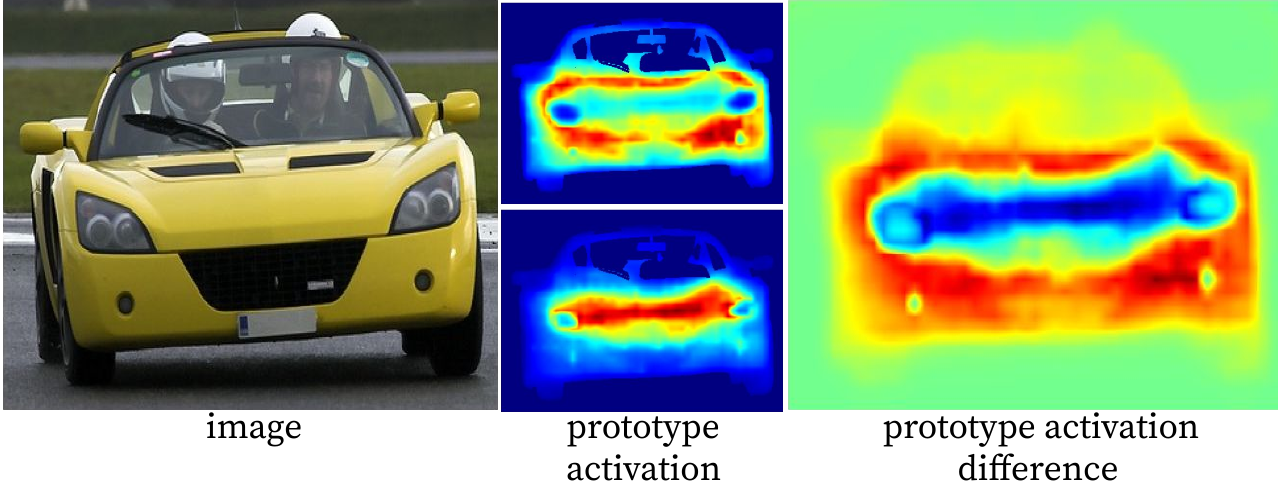}
\vspace*{-5mm}
\caption{Low value of $\lossarg{L}{J}$.}
\label{fig:loss_KLD_b}
\end{subfigure}
\caption{Comparison between high and low values of $\lossarg{L}{J}$ for the activation of two prototypes. $\lossarg{L}{J}$ has a high value if two prototypes of the same class activate in the same area (a). For this reason, we add $\lossarg{L}{J}$ as an additional component of the loss function to increase the variety of prototypes within each class (b).}
\label{fig:loss_KLD}
\end{figure}

\begin{figure}[t]
\begin{center}
\includegraphics[width=0.95\columnwidth]{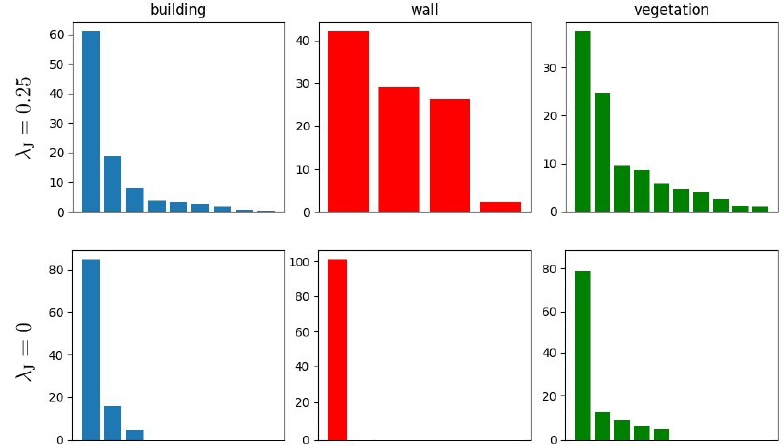}
\end{center}
\caption{
Histograms showing the assignment of feature map points to prototypes per class on Cityscapes. As the assignment, we understand finding the highest activated prototype for a given feature map point. In the top row, we present the model trained with the diversity loss $\lossarg{L}{J}$, while at the bottom without $\lossarg{L}{J}$. One can observe that the diversity loss increases the utilization of prototypes by the \our{}.}
\label{fig:histograms}
\end{figure}

We simplify the approach of~\cite{chen2019looks} and remove the \textit{cluster} and \textit{separation} losses, which we find redundant for our method. Instead, we combine the standard cross-entropy loss with an additional component which enforces same-class prototypes to be activated in different image areas (see \Cref{fig:loss_KLD_b}), resulting in optimal utilization of prototypes by the model (\Cref{fig:histograms}). In this section we will describe in detail how we construct the additional loss term that improves the diversity of prototypes.

\paragraph{Jeffrey's similarity}
Firstly, we start by introducing a function which will encourage the diversity of prototypes. Let us recall that 
{\em Jeffrey's divergence}~\cite{jeffreys1998theory} between two probability distributions $U$ and $V$
\begin{equation}
\D_J(U,V)=\tfrac{1}{2} \D_{KL}(U\|V)+\tfrac{1}{2}\D_{KL}(V\|U)
\end{equation}
is defined as the symmetrization of the Kullback-Leibler divergence. Clearly, $\D_J(U,V)=0$ iff $U=V$, and large value of $\D_J$ implies that the distributions concentrate on different regions. Now given a sequence of distributions
$U_1,\ldots,U_l$ we introduce their {\em Jeffrey's similarity}
by the formula
\begin{equation}
\mathcal{S}_J(U_1,\ldots,U_l)=\frac{1}{\binom{l}{2}}\sum_{i < j} \exp(-\D_J(U_i,U_j))
\end{equation}
Observe that Jeffrey's similarity is permutation invariant, it attains values in the interval $[0,1]$ and $\mathcal{S}_J(U_1,\ldots,U_n)=1$ iff $U_1=\ldots=U_n$. Moreover, if distributions $U_i$ have pairwise disjoint supports then $\mathcal{S}_J(U_1,\ldots,U_n)=0$.

\paragraph{Prototype-class-image distance vector}
Let $p\in\mathbf{P}_c $ be some prototype assigned to class $c\in C$ and $Z = f(x) \in \R^{H_d \times W_d \times D}$ be the feature map of some image $x$ after processing through the backbone $f$. Let also $Y_Z\in\R^{H_d \times W_d}$ be the ground-truth class labels per each feature map points of $Z$. We define \textit{prototype-class-image distance vector} between image feature map $Z$ and prototype $p$ as
\begin{equation}
v(Z, p) = \mathrm{softmax}(\norm{z_{ij}-p}^2 \, | \,  z_{ij}\in Z : Y_{ij} = c).
\end{equation}
The vector $v(Z, p)$ measures the relative activation of the prototype $p\in\mathbf{P}_c $ on parts of the image that are assigned to class $c$. We note that the length of the vector $v(Z, p)$ is equal to the number of points on the feature map in $Z$ assigned to class $c$.


\paragraph{Prototype diversity loss}
We define the \textit{prototype diversity loss} between the sequence of prototypes $\mathbf{P}_c=(p_1,\ldots, p_k) $ from the same class $c \in C$ on image feature map $Z$ as
\begin{equation}
\lossarg{L}{J}(Z,\mathbf{P}_c)  = \mathcal{S}_J(v(Z, p_1),\ldots, v(Z, p_k)).
\end{equation}
Note that $\lossarg{L}{J}(Z,\mathbf{P}_c)$ measures the difference between distributions of prototype activation within a sequence $\mathbf{P}_c$ on feature map points from $Z$ assigned to their class. It is minimized by lowering the Jeffrey's similarity of their prototype-class-image distance vectors. Finally, we introduce the \textit{total prototype diversity loss} for a feature map $Z$ and the set of all prototypes $\mathbf{P}$
\begin{equation}
\lossarg{L}{J} = \frac{1}{C} \sum\limits_{c=1}^C \lossarg{L}{J}(Z,\mathbf{P}_c). 
\end{equation}

The final loss during \textit{warmup} and \textit{joint training} is
\begin{equation}
\lossarg{L}{} = \lossarg{L}{CE} + \lossarg{\lambda}{J}\cdot \lossarg{L}{J},
\label{eq:joint_loss}
\end{equation}
where $\lossarg{L}{CE}$ is the cross entropy loss for pixel patch-wise classification and $\lossarg{\lambda}{J}$ as a hyperparameter that controls the weight of diversity of same-class prototypes within their assigned class.
Following the training protocol from~\cite{chen2019looks}, we add an additional L1-norm loss term on $w_h$ weights during the \textit{fine-tuning} phases, making the total loss in these phases equal to
\begin{equation}
\lossarg{L}{F} = \lossarg{L}{} + \lossarg{\lambda}{L1}\cdot \sum\limits_{c=1}^C \sum_{j:p_j \notin \boldsymbol{P}_c} | w_h^{(c, j)} |. 
\label{eq:finetuning_loss}
\end{equation}

\section{Experimental setup}
\label{sec:exp_set}

In all experiments, we use DeepLab~\cite{chen2017deeplab} with ResNet-101~\cite{he2016deep} weights pretrained either on ImageNet~\cite{deng2009imagenet} or COCO~\cite{lin2014microsoft}. We assign $10$ prototypes to each class, and set the prototype size to $D=64$. We set the weights of loss terms to $\lossarg{\lambda}{L1} = 10^{-4}$ and $\lossarg{\lambda}{J} \in \{0, 0.25\}$. For input images, we employ augmentation techniques, such as random cropping, horizontal flipping and scaling images by a factor in range $[0.5,1.5]$. We use batch size equal to $10$ and Adam~\cite{adam2014} optimizer with weight decay $5\cdot10^{-4}$ and $\beta_1=0.9$, $\beta_2=0.999$. We freeze the batch normalization parameters during training to avoid noisy normalization statistics due to small batch size. In the warmup phase, we use a constant learning rate of $2.5\cdot 10^{-4}$ and train for $3\cdot10^4$ steps. In the joint training phase, we start with learning rate of $2.5\cdot 10^{-5}$ for ResNet-101 weights and $2.5\cdot 10^{-4}$ for ASPP and prototype layers. We employ the polynomial learning rate policy~\cite{chen2017deeplab} with $power=0.9$, training for $3\cdot10^4$ steps. In both fine-tuning phases, we use a constant learning rate equal to $10^{-5}$ and train for $2000$ steps.

We run experiments on a single NVidia GeForce RTX 2080 GPU. For both datasets, the whole training procedure takes up to 48 hours. The code is written using PyTorch~\cite{paszke2019pytorch} and Pytorch Lightning~\cite{falcon2019pytorch} libraries.

\paragraph{Pascal VOC 2021.}
We evaluate \our{} on PASCAL VOC 2012 segmentation benchmark~\cite{pascal2012} that consists of $1464$ train, $1449$ validation, and $1446$ test images with pixel-level labels from $21$ distinct classes, including $20$ foreground classes and a background class. We use the offline augmented train\_aug dataset with $10582$ images provided in~\cite{pascal2012augmentation} for model training. However, we use the non-augmented training set for the prototype projection phase. We employ Multi-Scale inputs with max fusion (MSC)~\cite{chen2017deeplab} using scales $0.5$, $0.75$, and $1.0$. We set the image resolution to $321\times321$ pixels during training and evaluate on full images resized to $513\times513$ pixels for inference.

\paragraph{Cityscapes.} 
We also test \our{} on Cityscapes~\cite{cityscapes2016}, a large-scale image segmentation dataset that contains $2975$ train, $500$ validation, and $1525$ test images of street scenes. Following the suggestions of dataset authors, we train and evaluate the model on $19$ selected pixel classes and ignore the void class during training and evaluation. We use similar settings as those for PASCAL VOC 2012 with the following differences: lack of MSC following~\cite{chen2017deeplab}, training on random image crops of resolution $513\times513$ pixels, and evaluating on full images with the original resolution of $1024\times2048$ pixels.

%

%
%
%
%
%
%

\begin{table}[t]
\begin{center}
\begin{tabular}{lllcc}
\toprule
\multirow{2}{*}{Dataset} & \multirow{2}{*}{Method} & \multirow{2}{*}{Pretraining} &   \multicolumn{2}{c}{mIOU} \\
& & & val & test \\
\midrule
\multirow{4}{*}{Pascal} & DeepLabv2 & COCO & 77.69 & 79.70\\
\cmidrule{2-5} 
  & \our & COCO & 67.98 & 68.71 \\
\cmidrule{2-5}
  & \our & ImageNet & 72.05  & 72.92\\
\midrule
\multirow{4}{*}{Cityscapes} & DeepLabv2 & COCO & 71.40 & 70.40\\
\cmidrule{2-5}
 & \our & COCO & 55.35 & 56.77 \\
 \cmidrule{2-5}
 & \our & ImageNet & 67.23 & 67.04\\
\bottomrule
\end{tabular}
\end{center}
\caption{Performance of \our{} and the baseline method on the validation and test sets of PASCAL VOC 2012 and Cityscapes. The interpretability comes with a decrease in the mIOU. However, it is compensated with our diversity loss. }
\label{tab:results_val}
\end{table}


\begin{figure}[t]
\begin{center}
\includegraphics[width=0.8\columnwidth]{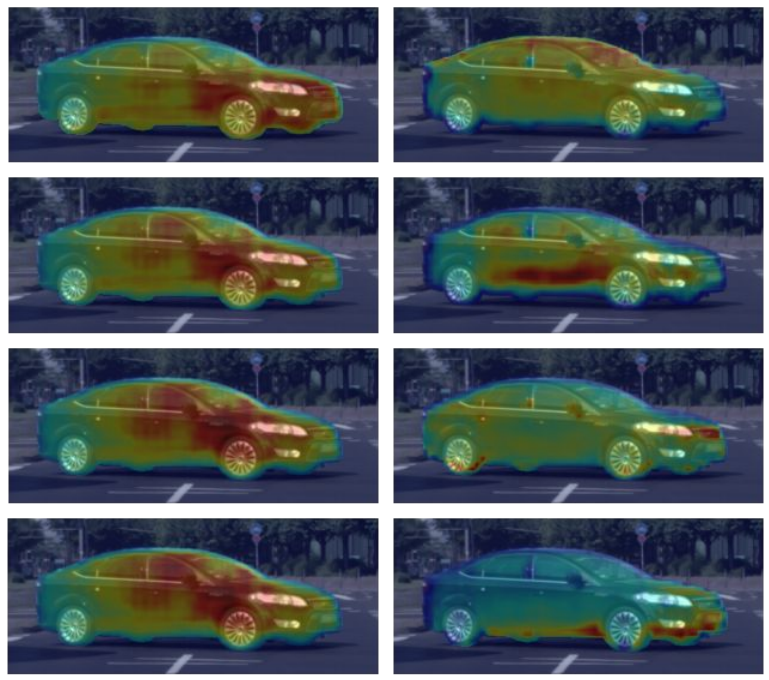}
\end{center}
\caption{Activation maps of four prototypes from class \textit{car} for \our{} trained without (left column) and with $\lossarg{L}{J}$ (right column). In the former, the activations of different prototypes overlap, in contrast to the latter, where the diversity of prototypes increases.}
\label{fig:kld_vs_no_kld_activation}
\end{figure}

\begin{table}[t]
\begin{center}
{\small{
\begin{tabular}{lccc}
\toprule
Dataset & $\lossarg{\lambda}{J}$ & prototype overlap (mIOU) & mIOU \\
\midrule
\multirow{2}{*}{PASCAL} & 0.00 & 48.16 & 69.60 \\
\cmidrule{2-4}
 & 0.25 & 26.59 & 72.05 \\
\midrule
\multirow{2}{*}{Cityscapes} & 0.00 & 57.99 & 61.60 \\
\cmidrule{2-4}
 & 0.25 & 24.09 & 67.23 \\
\bottomrule
\end{tabular}
}}
\end{center}
\caption{We analyze how the activation maps of two prototypes from the same class overlap each other. For this purpose, we binarize activation maps of all prototypes and calculate their mean IOU over all pairs of prototypes from the same class. The overlap is reduced by half after applying our $\lossarg{L}{J}$, which numerically confirms increased variability of prototypes. We also report the segmentation mIOU score in the rightmost column.}
\label{tab:kld_vs_no_kld_activation_miou}
\end{table}

\begin{figure*}[th]
\begin{center}
\centering
\includegraphics[width=0.9\linewidth]{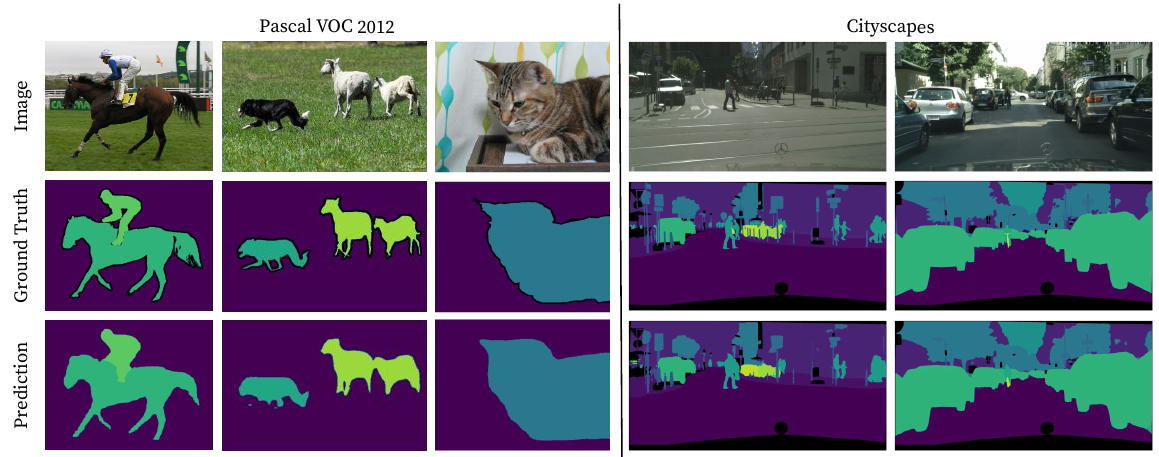}
\end{center}
\caption{Sample \our{} segmentations on PASCAL VOC 2012 (left) and Cityscapes (right). \our{} captures the overall object contours but may be inaccurate for fine-grained details. Note that pixels not considered in the evaluation are masked with black color in ground truth images.}
\label{fig:segmentation_examples}
\end{figure*}

\section{Results}
\label{sec:results}

\Cref{tab:results_val} presents the mean Intersection over Union (mIoU) scores obtained for the validation and test sets of PASCAL VOC 2012 and Cityscapes by the baseline method and \our{} with $\lossarg{\lambda}{J}=0.25$. We observe that the interpretability of \our{} comes with a decrease in mIOU compared to the baseline method. It could be caused by the constraint introduced by prototypes and, in our opinion, can be improved with more extended hyperparameter search. In the Supplementary Materials we present an ablation study on \our{}'s hyperparameters on Pascal VOC 2012, which shows the difficulty in bridging the gap to the baseline. \our{}, in contrast to~\cite{chen2017deeplab}, yields better results with weights obtained from the model pretrained on ImageNet classification than COCO segmentation task. We hypothesize that prototypes learned on ImageNet representation can be more informative because they correspond to a more generic task of image classification, whereas representation after pretraining on COCO segmentation can focus on more task-specific features such as object borders. Finally, we did not apply CRF~\cite{chen2017deeplab}, which can further improve the accuracy.

\paragraph{Influence of prototype diversity loss.} In \Cref{tab:kld_vs_no_kld_activation_miou} we compare the accuracy of \our{} with and without applying  $\lossarg{L}{J}$. We observe that \our{} achieves higher accuracy with $\lossarg{\lambda}{J}=0.25$ than $\lossarg{\lambda}{J}=0$. This could be attributed to the higher informativeness of diverse prototypes that leads to better generalization. To analyze this trend, we calculate an additional metric of prototypes overlapping, which we also present in  \Cref{tab:kld_vs_no_kld_activation_miou}. For this purpose, we binarize activation maps of all prototypes using $95$th percentile and calculate the mean IOU of highly activated regions over all pairs of prototypes from the same class. This overlap is reduced by half after applying our $\lossarg{L}{J}$. Hence, on average, for two prototypes from the same class, their highly activated regions have about 50\% overlap when $\lossarg{\lambda}{J}=0$ and only about 25\% when $\lossarg{\lambda}{J}=0.25$. \Cref{fig:kld_vs_no_kld_activation} presents activation maps of prototypes from class \textit{car} for models trained with $\lossarg{\lambda}{J} \in \{0, 0.25\}$ on Cityscapes with ImageNet pretraining. We observe that the prototypes of the model trained with non-zero $\lossarg{\lambda}{J}$ activate in semantically different regions, while the model with no diversity loss learns indistinguishable prototypes. To conclude, we observe that adding a non-zero $\lossarg{L}{J}$ increases the diversity of prototypes and allows for their interpretation as specific semantic object concepts.

\begin{table}[t]
\begin{center}
{\small{
\begin{tabular}{llcc}
\toprule
Dataset & Training stage & Num prototypes & mIOU \\
\midrule
\multirow{4}{*}{PASCAL} & warmup & 210 & 25.65 \\
\cmidrule{2-4}
& joint training  & 210 & 68.24 \\
\cmidrule{2-4}
& projection & 201 &  72.00 \\
\cmidrule{2-4}
& pruning &  133 &  72.05 \\
\midrule
\multirow{4}{*}{Cityscapes} & warmup & 190 & 31.45 \\
\cmidrule{2-4}
& joint training  & 190 & 65.38 \\
\cmidrule{2-4}
& projection & 188 & 67.24 \\
\cmidrule{2-4}
& pruning & 128 & 67.23 \\
\bottomrule
\end{tabular}
}}
\end{center}
\caption{Model performance after successive training stages for the validation set of PASCAL VOC 2012 and Cityscapes. The highest gain is achieved after joint training. However, the projection step is also beneficial. At the same time, pruning does not introduce significant performance improvement but removes around 30\% of irrelevant prototypes.}
\label{tab:ablation_stages_pascal}
\end{table}

\begin{figure*}[th]
\begin{center}
\includegraphics[width=0.9\linewidth]{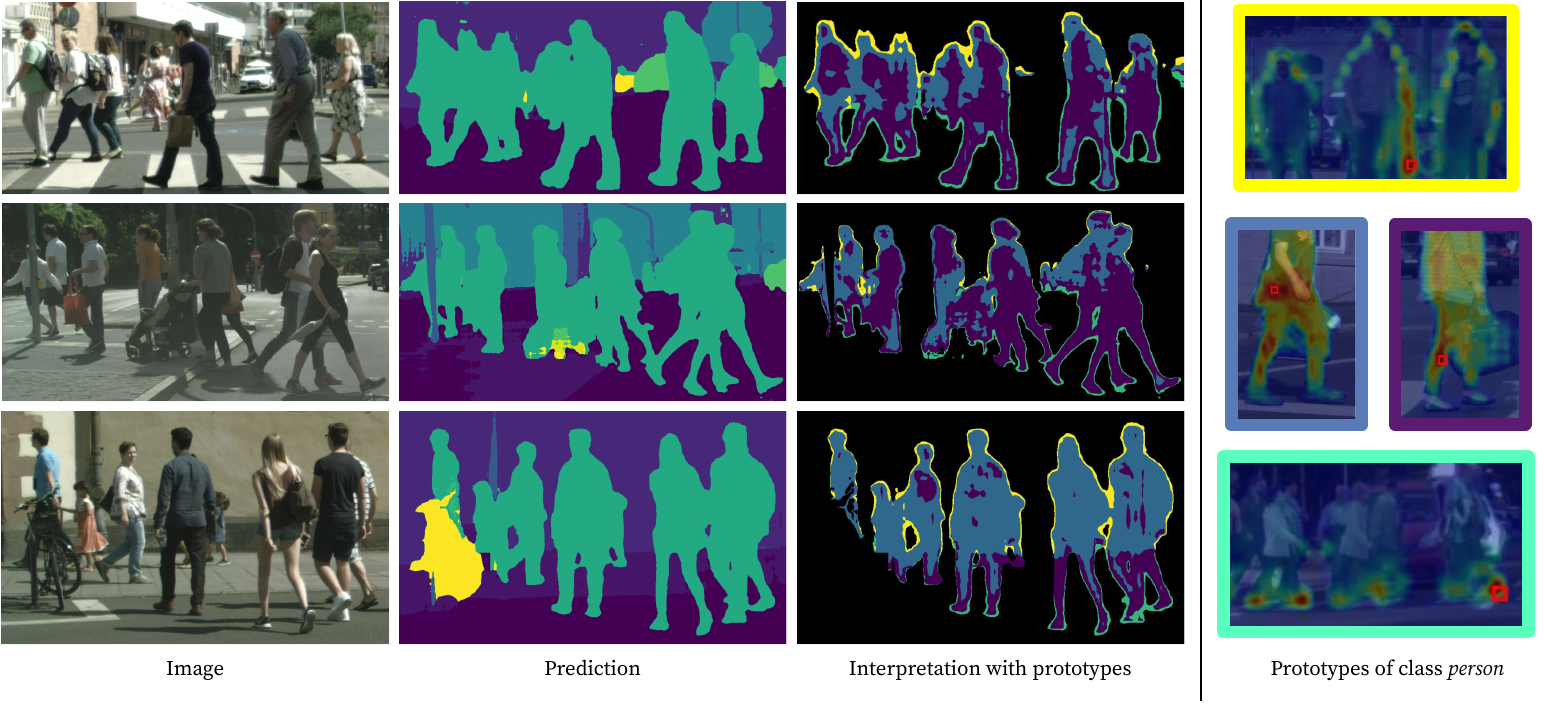}
\end{center}
\caption{Sample segmentation (second column) of images (first column) and interpretation with four prototypes (third column) from class \textit{person} obtained by \our{} on Cityscapes. Interpretation with prototypes is acquired by assigning the prototype with the maximal activation to a considered pixel. Pictures in the right column show the four prototypes (activating on legs, torso or other fragments), and their frame colors correspond to the colors from the third column.}
\label{fig:cityscapes_object_parts}
\end{figure*}

\paragraph{Segmentation with interpretable prototypes.} In \Cref{fig:segmentation_examples}, we present examples of segmentation maps predicted by the models pretrained on ImageNet with $\lossarg{\lambda}{J}=0.25$. Moreover, to exemplify that \our{} finds semantically meaningful prototypes, in \Cref{fig:prototype_object_parts_cats} we draw activation maps of prototypes from class \textit{cat} trained on PASCAL VOC 2012. We notice that the model learns prototypes representing the same semantic part concepts throughout different images. For instance, prototype 1 from~\Cref{fig:prototype_object_parts_cats} activates on cat's nose, and prototype 3 activates on outer rim of the cat's mouth. However, some prototypes can carry low-level information; for example, prototype 2 activates around the edge of the cat. Activations of prototypes can also be used to segment an image into semantic concepts, as shown in~\Cref{fig:cityscapes_object_parts}, where four prototypes of class \textit{person} are activated on different object fragments and could be interpreted as pointing to specific parts of an object, such as legs, torso, or boundary between the outline of a person and background. We provide more examples showing the interpretability of prototypes in the Supplementary Materials.

\paragraph{Accuracy after different training stages.} As described in \Cref{sec:multi_step_training}, \our{} employs a multi-stage procedure that affects model performance and the number of prototypes. In \Cref{tab:ablation_stages_pascal} we show mIOU scores of the models trained with ImageNet pretraining and $\lossarg{\lambda}{J}=0.25$ after different training phases, as well as the number of unique prototypes. We see that the model needs joint training of all layers, including the backbone, to achieve satisfactory accuracy. We also note that projection and pruning phases have no negative effect on model performance, even though they substantially reduce the number of prototypes.




\paragraph{\our{} with a different backbone.}

In this section, we show the adaptability of \our{} to a different backbone model. We extend U-Net~\cite{ronneberger2015u} with \our{} and evaluate it on the  EM segmentation challenge dataset from ISBI 2012~\cite{em_challenge}, which contains $30$ pixel-labeled microscopy images of \textit{Drosophila} larva. To perform the evaluation, we randomly divide the dataset into $20$ training and $10$ test samples. Our model achieves almost the same pixel error as U-Net (see ~\Cref{tab:unet}), while introducing model transparency with prototypes. In the Supplementary Materials, we provide the details about the training of the U-Net-based methods and some prediction examples 

\begin{table}[t]
\begin{center}
{\small{
\begin{tabular}{lcc}
\toprule
Model & mIOU & Pixel error \\
\midrule
U-Net & 78.74 & 0.0537 \\
\our{} (U-Net backbone) & 76.58 & 0.0540  \\
\bottomrule
\end{tabular}
}}
\end{center}
\caption{Comparison between baseline U-Net and \our{} with U-Net backbone on the EM segmentation challenge dataset. \our{} achieves pixel error comparable to the baseline model while introducing interpretability of its predictions.}
\label{tab:unet}
\end{table}

\section{Conclusions}
\label{sec:conc}

In this work, we presented \our{}, a model for semantic segmentation that constructs its decisions by referring to prototypes found on the training set. Moreover, to increase the variability of prototypes within each class, we provide a novel diversity loss function. As presented in experiments conducted on various semantic segmentation datasets, we developed a method that allows for interpretation of obtained segmentation and achieves accuracy comparable to the baseline approaches.

The possible areas of future work include enhancing the precision of \our{} and applying it on novel state-of-the-art segmentation architectures or more challenging segmentation tasks. We also see room for improvement in better prototype selection or sharing prototypes between classes.

\paragraph{Code availability.} We made the code available at: \href{https://github.com/gmum/proto-segmentation}{https://github.com/gmum/proto-segmentation}

\section*{Acknowledgements}

The research of D. Rymarczyk was carried out within the research project ``Bio-inspired artificial neural network'' (grant no. POIR.04.04.00-00-14DE/18-00) within the Team-Net program of the Foundation for Polish Science co-financed by the European Union under the European Regional Development Fund.
The work of M. Sacha, J. Tabor and B. Zieliński is supported by the National Centre of Science (Poland) Grant No. 2021/41/B/ST6/01370, and the work of Ł. Struski is supported by by the National Centre of Science (Poland) Grant No. 2020/39/D/ST6/01332.

\clearpage

{\small
\bibliographystyle{ieee_fullname}
\bibliography{egbib}
}

\end{document}